\documentclass[letterpaper, 10 pt, conference]{ieeeconf}  

\usepackage{amsmath,amssymb,breqn}
\usepackage[ruled,vlined]{algorithm2e}
\usepackage{lipsum}
\usepackage[normalem]{ulem}
\useunder{\uline}{\ul}{}
\usepackage{makecell}\usepackage{color}
\usepackage{subfigure}
\usepackage{graphicx}
\usepackage[    colorlinks=true,
    linkcolor=blue,
    filecolor=blue,      
    urlcolor=blue,
    citecolor=blue,]{hyperref}

\IEEEoverridecommandlockouts                              

\title{\LARGE \bf
Unguided Self-exploration in Narrow Spaces with Safety Region Enhanced Reinforcement Learning for Ackermann-steering Robots
}

\author{Zhaofeng Tian$^{1}$, Zichuan Liu$^{2}$, Xingyu Zhou$^{3}$ and Weisong Shi$^{1}$  
\thanks{}
\thanks{$^{1}$The CAR Lab, University of Delaware, Newark, USA
        {\tt\small \{zhaofeng,weisong\}@udel.edu}}%
\thanks{$^{2}$Department of Computer Science, Wayne State University,
        Detroit, USA
        {\tt\small zichuanliu@wayne.edu}}%
\thanks{$^{3}$Department of Electrical and Computer Engineering, Wayne State University,
        Detroit, USA
        {\tt\small xingyu.zhou@wayne.edu}}%
}

\begin{document}

\maketitle
\thispagestyle{empty}
\pagestyle{empty}

\begin{abstract}
In narrow spaces, motion planning based on the traditional hierarchical autonomous system could cause collisions due to mapping, localization, and control noises, especially for car-like Ackermann-steering robots which suffer from non-convex and non-holonomic kinematics. To tackle these problems, we leverage deep reinforcement learning which is verified to be effective in self-decision-making, to self-explore in narrow spaces without a given map and destination while avoiding collisions. Specifically, based on our Ackermann-steering rectangular-shaped ZebraT robot and its Gazebo simulator, we propose the rectangular safety region to represent states and detect collisions for rectangular-shaped robots, and a carefully crafted reward function for reinforcement learning that does not require the waypoint guidance. For validation, the robot was first trained in a simulated narrow track. Then, the well-trained model was transferred to other simulation tracks and could outperform other traditional methods including classical and learning methods. Finally, the trained model is demonstrated in the real world with our ZebraT robot.  \url{https://sites.google.com/view/rl4exploration}
\end{abstract}



\section{INTRODUCTION}

Autonomous driving and robotics technologies have undergone significant advancements in recent times. A notable emerging application of these technologies is in hazardous environment operations, where robots can be utilized to perform essential tasks that would otherwise put human lives at risk. Examples of such applications include underground mining, tunnel/pipeline inspection, and post-disaster search~\cite{mining,post}. In these scenarios, in addition to the inherent risk to human life, a key challenge is the need for autonomous robots to self-explore narrow spaces while avoiding collisions. This task is particularly arduous for car-like Ackermann-steering robots for two reasons. First, such robots are subject to non-convex kinematic constraints~\cite{libai1}, e.g., the robot cannot drive sideways or turn in place, which limits its maneuverability. Second, these robots typically have a rectangular shape, introducing challenges to define free space for collision avoidance in geometric methods~\cite{libai2}.


One straightforward approach for the aforementioned application is to adopt a traditional hierarchical methodology, which involves a sequence of tasks such as sensing, perception, planning, and control. Nevertheless, this method faces several significant limitations that render it insufficient for the current tasks. First, planning and control necessitate a premade map, which is unavailable in the self-exploration context. Second, even with a premade map, obstacle parameterization, and non-convex obstacle convexification can pose significant challenges. In general, obstacle convexification based on the decomposition and dilation of the original premade map can limit the accessible areas and result in planning failures, particularly in narrow spaces where the obstacle is non-convex and curvy. Third, the dynamic uncertainty introduced by stacked mapping, localization, and control errors cannot be overlooked.


Reinforcement Learning (RL), particularly Deep Reinforcement Learning (DRL) -- which employs deep neural networks for function approximations or representations -- has recently achieved significant success across a diverse range of applications. These include Atari video games~\cite{DQN}, Alpha Go~\cite{Go}, and real-world scenarios like robot manipulators~\cite{arm}, navigation~\cite{trainning_wheels}, and transportation~\cite{its}. The fundamental concept behind RL involves sequential decision-making in unknown environments where the agent learns to act by interacting with the environment through trial-and-error to maximize the long-term reward. Furthermore, the use of deep neural networks in DRL allows the agent to handle large-scale states and learn the optimal policy directly from raw inputs without hand-engineered features or domain heuristics. To harness the potential of DRL, various learning algorithms have been developed in recent years, including Deep Q-Network (DQN) for discrete action space and Deep Deterministic Policy Gradient(DDPG)~\cite{DDPG}, Soft Actor-Critic (SAC)~\cite{SAC}, and Proximal Policy Optimization (PPO)~\cite{PPO} for continuous action space.


In this paper, we present the first step towards applying Deep Reinforcement Learning (DRL) to address the problem of self-exploration in narrow spaces without using pre-made maps of the environments. Specifically, we focus on the scenario where a rectangular-shaped Ackermann-steering robot aims to efficiently explore general narrow spaces without any guidance from waypoints or destinations, while avoiding collisions. By narrow spaces, we refer to non-branching passages that constrain an Ackermann-steering robot from turning around for 180 degrees without backing up or the space that is narrower than the circumscribed circle of the robot body when viewed from above.  Two unique challenges arise in our case: (i) detecting collisions in narrow spaces is a challenging task when the robot is rectangular-shaped and uses a Lidar or laser scanner to sense the environment. The standard approach of setting a scan range value to cover the robot body can lead to an over-coverage problem, which, in turn, can result in inefficient exploration; (ii) as with many DRL applications, designing a suitable reward function for the agent is a critical challenge. In our scenario, motivating the car-like robot to explore without guidance from a destination or waypoint and designing a steering and reversing mechanism for its non-convex kinematics to avoid collisions are not addressed in the previous traditional way-point guided DRL reward function for differential robots. This requirement necessitates more delicate reward shaping to balance various working conditions.



\textbf{Our contributions.} We make two main contributions to address the challenges outlined above. First, we introduce a rectangular safety region that is tailored for rectangular-shaped robots to accurately represent robot states and enable precise collision detection. Second, we carefully design a reward function that balances exploration and collision avoidance while without relying on waypoint guidance for narrow spaces. To systematically evaluate our proposed methods, we conduct a four-step procedure as follows:


(i) \emph{Collision tests:} We compare our proposed safety region representation for rectangular-shaped robots with other laser-scan-based representations for collision detection. The results show that our safety region can detect collisions more accurately with the same number of discretized laser scans.


(ii) \emph{Training in simulation:} We incorporate our proposed reward function into five mainstream DRL algorithms - DDPG, DQN, PPO, PPO-discrete, and SAC - and benchmark their performance in a simulated narrow track. Using the selected DDPG algorithm, we then train multiple models with other reward functions for contrastive and ablation evaluation studies. We use multi-seeds for each algorithm and model, and each seed is episode-fixed for fairness.


(iii) \emph{Sim-to-sim evaluations:} We transfer the different trained models from step (ii) to seen and unseen scenarios and evaluate their performance by success rate, fail rate, collision rate, and time cost. We compare the proposed reward function with a traditional waypoint-guided reward function (contrastive model) and models with different reward components (ablation models). We also benchmark other classical map-based geometry methods and the Imitation Learning (IL) method. The model using the proposed reward function demonstrates a convincing generalization ability and outperforms other methods in most scenarios.


(iv) \emph{Sim-to-real demonstration:} We transfer the learned policy from step (ii) directly to our real-world ZebraT robot and test it on three real-world narrow tracks. The trained model with the proposed reward function successfully completes all three tracks without collisions. This not only validates the effectiveness of our proposed methods but also demonstrates the generalization power of the learned policy.

\section{RELATED WORK}

\subsection{Classical Motion Planning}
Traditional map-based methods of motion planning have dominated the area for a long time, which are built upon mapping, global planning, and local planning workflow. Global planning primarily refers to pathfinding problems, where A*~\cite{astar} and Dijkstra~\cite{dijkstra} algorithms are two classical methods used in Robot Operating System (ROS)~\cite{ros} navigation stack. Whereas local planning pays more attention to collision avoidance and kinodynamic feasibility. Dynamic Window Approach (DWA)~\cite{dwa} and Timed Elastic Band (TEB)~\cite{teb} planners are two local motion planners  integrated with ROS and have been widely used in the community, where DWA is suitable for differential drive robots, and TEB planner is able to consider the kinematic constraints of a car-like robot. And some motion planning methods aim to solve collision avoidance problems in a dynamic environment, e.g., Velocity Obstacle (VO) method~\cite{vo}, and its derivatives Reciprocal Velocity Obstacle (RVO)~\cite{rvo}, Optimal Reciprocal Collision Avoidance (ORCA) methods~\cite{orca}.

\subsection{DRL Motion Planning}
A navigation problem for a mobile robot is usually defined as how to move from point A to point B. Machine learning-based motion methods are promising to solve navigation problems in an unknown environment. \cite{il} practices end-to-end navigation through IL, where the model is trained with human actions in a supervised paradigm. However, this supervised learning method is highly constrained by human policy and not guaranteed when validating set changed~\cite{trainning_wheels}. Plenty of research~\cite{curiosity,discrete,autorl,unknown_dynamic,robust} has demonstrated DRL is suitable to generate motions for robot navigation problems, where the difference between the current position and the destination is taken as a reward to facilitate the robot approaching the destination. Meanwhile, research~\cite{motion_dynamic,DWA_RL,moe} studies obstacle avoidance among dynamic obstacles using DRL methods. However, all these studies in subsections A and B, do not scope in narrow spaces and the robot models are round or near-round-shaped and use differential driving. While for a rectangular Ackermann-steering robot, the fixed range detection and the policy trained in open spaces would be insufficient. The other challenge for the Ackermann-steering robot is its non-convex action space, hence robot may need to back up for heading angle adjustments, so it needs to explore actions with negative linear velocities whereas a differential robot does not, which induces more uncertainties for learning convergency. 

\subsection{Narrow Space Motion Problem}
Moving a ``piano" in a narrow space without collision is a challenging task~\cite{mover}. A collision-free path sometimes is unachievable under some circumstances when using obstacle dilation on a map. Research ~\cite{narrow} used a Lyapunov-based control strategy to move a car-like robot in a narrow lane. However, this method costs daunting computations and requires a parameterized environment that is not available in exploration scenarios.

\begin{table*}[htb!]
\caption{Related work benchmark}
\label{related}
\centering
\scalebox{0.80}{
\begin{tabular}{lllllllllllll}

\hline
{\textbf{Article}} & { \textbf{Destination}} & { \textbf{Localization}} & \makecell[c]{\textbf{Premade}\\ \textbf{Map}}  & \makecell[c]{\textbf{Narrow}\\ \textbf{Space}}   & \makecell[c]{\textbf{Backup}\\ \textbf{Allowed}}  &\makecell[c]{\textbf{Collision}\\ \textbf{Detection}}  & \makecell[c]{\textbf{Robot}\\ \textbf{Shape}}  &\makecell[c]{\textbf{Robot}\\ \textbf{Kinematic}}   & \makecell[c]{\textbf{RL Algorithm}\\ \textbf{Benchmark}}  & \makecell[c]{\textbf{Sim-to}\\ \textbf{-real}}  &\makecell[c]{\textbf{Raw}\\ \textbf{Input}}   &\makecell[c]{\textbf{Action}\\ \textbf{Output}}  \\ \hline

               \makecell[c]{\cite{bnd}} & \makecell[c]{N} & \makecell[c]{N}  &\makecell[c]{N}  & \makecell[c]{N}  & \makecell[c]{N}  & \makecell[c]{-}  & \makecell[c]{close to\\ round} & \makecell[c]{differential}  &\makecell[c]{DQN}   & \makecell[c]{Y} &\makecell[c]{RGB }  & \makecell[c]{v,w} \\  
                \makecell[c]{\cite{successor_features}} &  \makecell[c]{N}   & \makecell[c]{N} & \makecell[c]{N} & \makecell[c]{N} & \makecell[c]{N} & \makecell[c]{-} & \makecell[c]{round} & \makecell[c]{differential} & \makecell[c]{DQN} & \makecell[c]{Y} & \makecell[c]{RGBD} & \makecell[c]{d,theta} \\
               \makecell[c]{\cite{target}} & \makecell[c]{Y} & \makecell[c]{N} & \makecell[c]{N} & \makecell[c]{N} & \makecell[c]{N} & \makecell[c]{N} & \makecell[c]{round} & \makecell[c]{differential} & \makecell[c]{A3C} & \makecell[c]{Y} & \makecell[c]{RGB} & \makecell[c]{d,theta} \\
               \makecell[c]{\cite{towards_cognitive}} & \makecell[c]{N} & \makecell[c]{N} & \makecell[c]{N} & \makecell[c]{N} & \makecell[c]{N} & \makecell[c]{fixed RGBD\\ camera depth} & \makecell[c]{round} & \makecell[c]{differential} & \makecell[c]{DQN} & \makecell[c]{Y} & \makecell[c]{RGBD} & \makecell[c]{v,w} \\
               \makecell[c]{\cite{towards_monocular}} & \makecell[c]{N} & \makecell[c]{N} & \makecell[c]{N} & \makecell[c]{N} & \makecell[c]{N} & \makecell[c]{N} & \makecell[c]{round} & \makecell[c]{differential} & \makecell[c]{DQN} & \makecell[c]{Y} & \makecell[c]{RGBD} & \makecell[c]{v,w}\\
               \makecell[c]{\cite{moe}} & \makecell[c]{Y} & \makecell[c]{2D lidar} & \makecell[c]{Y} & \makecell[c]{N} & \makecell[c]{N} & \makecell[c]{fixed laser range} & \makecell[c]{round} & \makecell[c]{differential} & \makecell[c]{PPO} & \makecell[c]{N} & \makecell[c]{2D laser\\ scans} & \makecell[c]{v,w} \\
                \makecell[c]{\cite{discrete}} & \makecell[c]{Y} & \makecell[c]{2D lidar} & \makecell[c]{Y} & \makecell[c]{N} & \makecell[c]{N} & \makecell[c]{fixed laser\\ range} & \makecell[c]{round} & \makecell[c]{differential} & \makecell[c]{DQN/DDPG/PPO} & \makecell[c]{Y} & \makecell[c]{2D laser\\ scans} & \makecell[c]{v,w} \\
                \makecell[c]{\cite{autorl}} &  \makecell[c]{Y} & \makecell[c]{2D lidar} & \makecell[c]{Y} & \makecell[c]{N} & \makecell[c]{N} & \makecell[c]{fixed laser\\ range} & \makecell[c]{round} & \makecell[c]{differential} & \makecell[c]{AutoRL} & \makecell[c]{Y} & \makecell[c]{2D Laser\\ scans} & \makecell[c]{v,w} \\
                \makecell[c]{\cite{unknown_dynamic}} & \makecell[c]{Y} & \makecell[c]{simulation} & \makecell[c]{Y} & \makecell[c]{N} & \makecell[c]{N} & \makecell[c]{simulation} & \makecell[c]{dot} & \makecell[c]{differential} & \makecell[c]{A3C} & \makecell[c]{N} & \makecell[c]{N} & \makecell[c]{v,w} \\
               \makecell[c]{\cite{trainning_wheels}} & \makecell[c]{Y} & \makecell[c]{2D lidar} & \makecell[c]{Y} & \makecell[c]{N} & \makecell[c]{N} & \makecell[c]{fixed laser\\ range} & \makecell[c]{close to\\ round} & \makecell[c]{differential} & \makecell[c]{DDPG} & \makecell[c]{Y} & \makecell[c]{2D laser\\ scans} & \makecell[c]{v,w} \\
                \makecell[c]{\cite{robust}}& \makecell[c]{Y} & \makecell[c]{2D lidar} & \makecell[c]{Y} & \makecell[c]{N} & \makecell[c]{N} & \makecell[c]{fixed laser\\ range} & \makecell[c]{round} & \makecell[c]{differential} & \makecell[c]{DDPG} & \makecell[c]{Y} & \makecell[c]{2D Laser\\ scans} & \makecell[c]{v,w} \\
                \makecell[c]{\cite{virtual_to_real}}& \makecell[c]{Y} & \makecell[c]{2D lidar} & \makecell[c]{Y} & \makecell[c]{N} & \makecell[c]{N} & \makecell[c]{fixed laser\\range} & \makecell[c]{round} & \makecell[c]{differential} & \makecell[c]{DDPG} & \makecell[c]{Y} & \makecell[c]{2D Laser\\scans} & \makecell[c]{v,w} \\
               \makecell[c]{\cite{motion_dynamic}} & \makecell[c]{N} & \makecell[c]{N} & \makecell[c]{N} & \makecell[c]{N} & \makecell[c]{N} & \makecell[c]{fixed laser\\range} & \makecell[c]{close to\\round} & \makecell[c]{differential} & \makecell[c]{A3C} & \makecell[c]{Y} & \makecell[c]{2D Laser\\scans} & \makecell[c]{v,w} \\
                \makecell[c]{\cite{DWA_RL}}& \makecell[c]{Y} & \makecell[c]{odometry} & \makecell[c]{N} & \makecell[c]{N} & \makecell[c]{N} & \makecell[c]{fixed laser\\range} & \makecell[c]{round} & \makecell[c]{differential} & \makecell[c]{PPO} & \makecell[c]{Y} & \makecell[c]{2D Laser\\scans} & \makecell[c]{v,w} \\
               \makecell[c]{Our} & \makecell[c]{N} & \makecell[c]{N} & \makecell[c]{N} & \makecell[c]{\color{red}Y} & \makecell[c]{\color{red}Y} & \makecell[c]{\color{red}{tunable safety} \\ \color{red}region }  & \makecell[c]{\color{red}rectangular\\ \color{red}(car like)} & \makecell[c]{\color{red}Ackermann\\ \color{red}steering} & \makecell[c]{DQN/DDPG\\PPO/SAC\\/PPPO-discrete} & \makecell[c]{Y} & \makecell[c]{\color{red}3D lidar\\ \color{red}point clouds} & \makecell[c]{v,w} \\ \hline
               
\end{tabular}

}
\end{table*}

\subsection{Innovation of This Study }
A comparison between related works with this study is listed in Table~\ref{related}. This study considers narrow space and validated with a rectangular-shaped Ackermann-steering car-like robot with 3D Lidar mounted, which is different from previous studies that are based on open space and smaller round-shaped differential robots. Additionally, this study uses a tunable safe region to detect a collision instead of a fixed range value. Moreover, no map and destination information is needed in this study, and backing up is allowed to avoid the obstacles straight ahead.

\section{PROBLEM FORMULATION AND BACKGROUND}

We model the narrow space self-exploration problem as a Markov Decision Process (MDP). In an MDP, an agent (i.e., the robot) interacts with the environment through a sequence of states, actions, and rewards. At each time step $t\in [0,T]$, the agent observes the current state $s_t$, executes an action $a_t$, obtains a reward $r_t$, and transitions to next state $s_{t+1}$. The goal is to maximize the cumulated reward from step $t$ onward, which is denoted by $G_t = \sum_{k=t}^{T} \gamma^{k-t} r_k$, where $\gamma $ is a factor to discount the future reward. In this work, we benchmark several DRL algorithms to tackle the MDP problem, wherein DQN and DDPG are two typical algorithms that respectively work on discrete and continuous action space.

\subsection{Deep Q Networks (DQN)}
Combining the original Q-learning with deep neural networks, DQN approximates an action-value function in a Q-earning framework with a neural network. The optimal action-value function $Q^*(s_t,a_t) = \max_\pi\mathbb{E}[G_t|s_t,a_t,\pi]$ is defined as the maximum expected return w.r.t. the state $s_t$ and the action $a_t$, where $\pi$ is a policy that maps states to actions. The optimal action-value function is subject to the Bellman equation so that it can be written as equation~(1).
\begin{equation}
Q^*(s_t,a_t) = \mathbb{E}_{s_{t+1}}[r_t+\gamma \max\limits_{a} Q^*(s_{t+1},a) | s_t, a_t]
\end{equation}

Then by using the value iteration equation $Q_{t+1}(s_t,a_t) = \mathbb{E}_{s_{t+1}}[r_t+\gamma\max_{a_{t+1}}Q_t(s_{t+1},a_{t+1})|s_t,a_t]$, the $Q_t$ can converge to $Q^*$ as time step $t\rightarrow\infty$. In the practice of DQN, the $Q_t$ value can be predicted by a deep neural network parameterized by $\theta$, and the network is trained with the loss function $L(\theta_t) = \mathbb{E}_{s_t,a_t}[(y_t - Q(s_t,a_t,\theta_t))^2]$, where  $y_t = \mathbb{E}_{s_{t+1}}[r_t + \gamma\max_{a}Q(s_{t+1},a_{t+1},\theta_{t-1})]$ is the target for the iteration at $t$.


\subsection{Deep Deterministic Policy Gradient (DDPG)}
Derived from the actor-critic model~\cite{AC} and policy gradient method~\cite{PG}, DDPG not only estimates the $Q$ value with a critic network parameterized by $\theta^Q$ but also estimates optimal actions with an actor network parameterized by $\theta^\pi$. Wherein the critic network is updated with a method similar to DQN that uses a loss function, while the actor network is updated with a policy gradient:
\begin{dmath}
\nabla _{\theta^\pi}J \approx \mathbb{E}[\nabla _{\theta^\pi}Q(s,a|\theta^Q)|_{s=s_t, a = \pi(s_t|\theta^\pi)}] =
\mathbb{E}[\nabla _{a}Q(s,a|\theta^Q)|_{s=s_t, a=\pi(s_t)}\nabla _{\theta^\pi}\pi(s|\theta^\pi)|_{s=s_t}]
\end{dmath}
where $J=\mathbb{E}_{r_t,s_t,a_t}[G_t]$ is the expected return from the beginning state which is subject to the start distribution, $Q(s,a|\theta^\pi)$ and $a = \pi(s_t|\theta^\pi)$ respectively denote the action value w.r.t. the critic network and the policy function w.r.t. the actor network. Framed in such an actor-critic structure, DDPG is suitable to compute an action for continuous control problems with high efficiency.


\section{Methods}

In this section, we present our main methods to apply DRL to solve the problem of narrow space self-exploration. To this end, we first develop a novel state representation method tailored for rectangular-shaped robots, which in turn helps to detect collisions in narrow spaces. Further, we carefully design a reward function with various components to trade off specific task requirements.

\subsection{State Representation with Rectangular Safety Region}

\begin{figure}[htbp]
\centering
\subfigure[Fixed range]{
\begin{minipage}[t]{0.5\linewidth}
\centering
\includegraphics[width=1.93in]{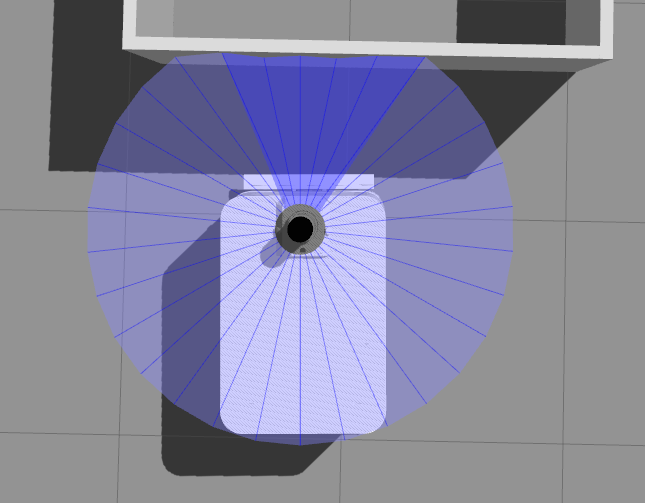}
\end{minipage}%
}%
\subfigure[Safety region]{
\begin{minipage}[t]{0.5\linewidth}
\centering
\includegraphics[width=1.5in]{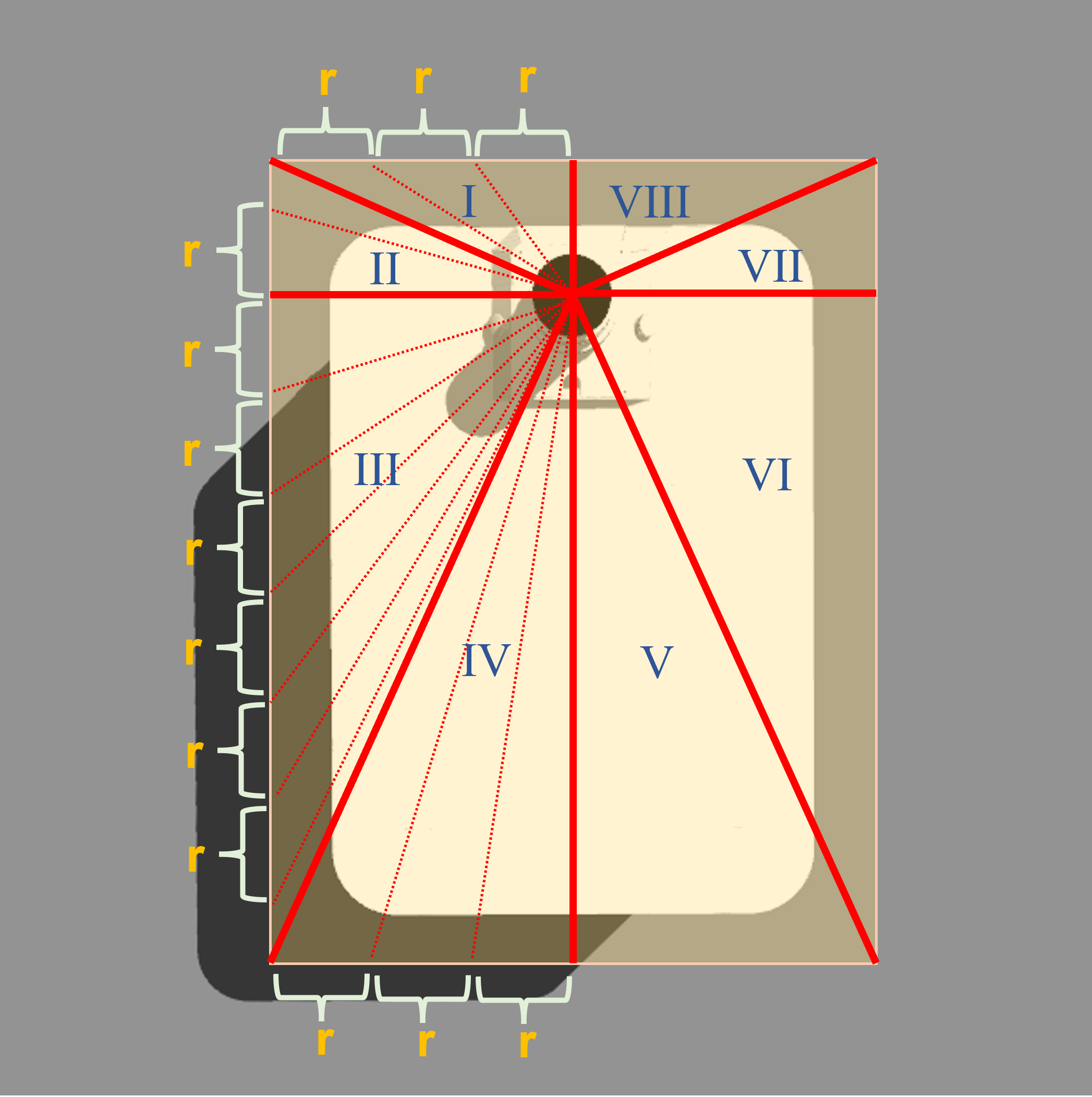}
\end{minipage}%
}%
\centering
\caption{Collision detection, using the fixed range, a circular area will cause over coverage. Instead, the proposed rectangular safety region method would detect collisions around the rectangular area.}
\label{state}
\end{figure}

As mentioned in the ``piano mover'' paper~\cite{mover}, it is a complex problem to figure out whether the ``piano" will collide with a curved wall or not. In most previous research, the robot is round-shaped, the state of which is represented by discretized laser scans with a fixed angle interval~\cite{discrete, trainning_wheels}. Consequently, collision detection can be implemented simply with a fixed range value. However, for a rectangular-shaped robot, this method could cause over-coverage problems shown in Fig.~\ref{state}. Since the Lidar is not usually mounted at the robot center, such that using a fixed range to cover the robot may generate a larger area that can be detected as collisions by mistakes, which subsequently makes more spaces inaccessible for the robot and detriments narrow space performance.


\begin{figure*}[h] 
\centering 
\includegraphics[width=0.7\textwidth]{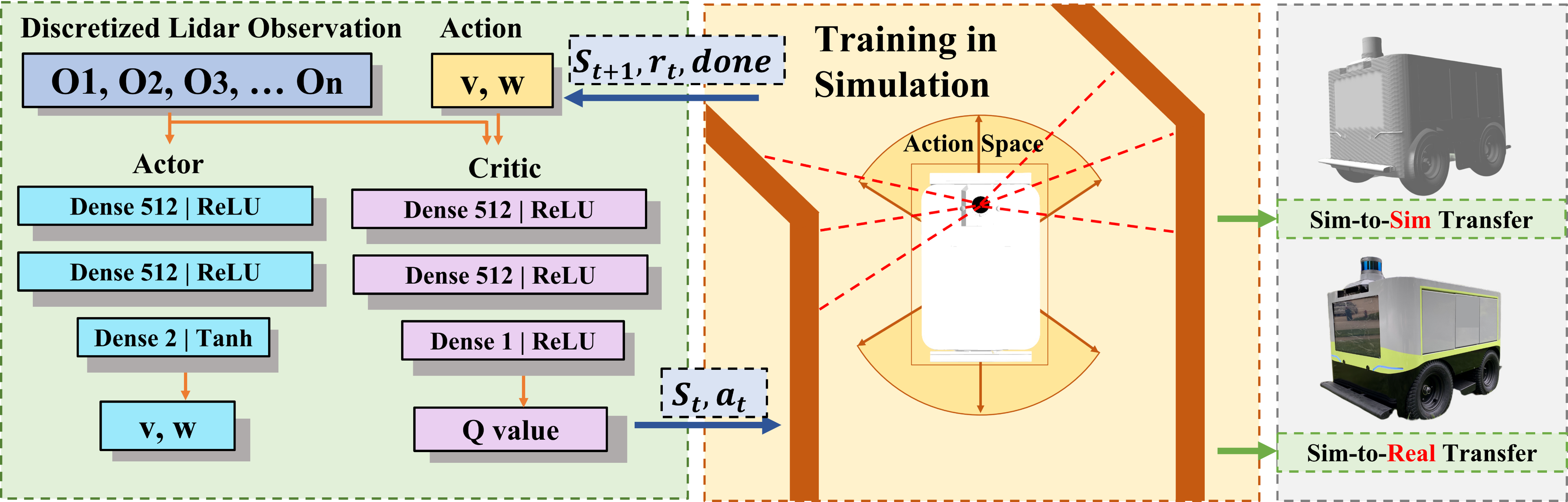} 
\caption{Workflow of reinforcement learning, the shown networks are used in DDPG training.} 
\label{workflow} 
\end{figure*}

\begin{algorithm}[!ht]
\DontPrintSemicolon
\SetKwInput{KwInput}{Input}                
\SetKwInput{KwOutput}{Output}              
\SetKwFunction{FDiscretize}{Discretize}
\SetKwFunction{FMain}{Main}
\SetKwFunction{FCollision}{Collision}
  \KwInput{$laserScans$, resolution $r$}
  \KwOutput{$V_{index}$, $V_{range}$, $V_{obs}$, $boolCollision$}
  \SetKwProg{Fn}{Def}{:}{}
  \;
\Fn{\FDiscretize{$laserScans$, $r$}}
{
    $inc\leftarrow$ $ round(2*pi/  n_{scans})$ \;
    \For{$phase \leftarrow 1$ \KwTo $8$}
    {   
        $n_{scans} \leftarrow l_{side} / r  $\;
        \For{$i_{scan} \leftarrow 0$ \KwTo $n_{scans}$}
        {
            $angle \leftarrow $ angle between 0 and the scan\;
            $ range \leftarrow $ range of the $i_{scan}$ \;
            $V_{index}$.append ($angle / inc$)\;
            $V_{range}$.append ($range$)\;
        }
    }
    \For{$i \leftarrow 0$ \KwTo len($V_{index}$)}
    {
        $V_{obs}$.append($laserScans[V_{index}[i]]$)
    }
    \KwRet{$V_{index}, V_{range}, V_{obs}$}

}
  \;
\Fn{\FCollision{$laserScans$}}
{
    $V_{obs}, V_{range}\xleftarrow[]{}$ []\;
    $boolCollision\xleftarrow[]{}$ $false$\;
    $V_{obs}, V_{range}\xleftarrow[]{}$ Discretize($laserScans$)\;
    
  \For{$i \leftarrow 0$ \KwTo len($V_{index}$)}
  {
    \If{$V_{obs}[i] \leq V_{range}[i]$}
    {
    $boolCollision\xleftarrow[]{}$ $true$\;
    \KwRet $boolCollision$\;
    }

  }
}

\caption{Rectangular Safety Region}
\end{algorithm}

To detect collisions more accurately, as expressed in Algorithm 1, we propose a rectangular-shaped safety region to represent the robot state. Also shown in Fig.~\ref{state}(b), the laser scans now are not discretized by a fixed interval angle anymore, instead, the angle intervals become adaptive to the robot's rectangular-shaped body so that the discretized scans could evenly cover the four sides of the rectangular safety region (yellow rectangular) with a resolution value $r$, which means for every side, the discretized scans could fall into it with an interval distance $r$. Besides, the scans in the directions of some important axis and diagonals (bold red lines) are selected, which split the rectangular safety region into 8 phases. Then in every phase, the indices~$V_{index}$ and the collision ranges~$V_{range}$ of the discretized scans can be calculated to detect collisions. And further, we use the ranges (the distance or length of each ray) of the discretized scans $V_{obs}$ to represent the robot state.


\subsection{Reward Function Shaping}
One key that distinguishes our reward function is that we do not take the destination into account because, in our considered scenario, a car-like robot needs to explore a narrow space without a certain destination while obeying its Ackermann-steering constraints. In contrast, the destination information in the previous papers is taken into reward function shaping, which could help the model convergence and training efficiency. { Hence how we urge the robot to move forward towards an ``open'' direction without a destination, instead of dawdling or going back, becomes a question. In this case, we craft a reward function comprises of four components, $R_f$, $R_o$, $R_m$, and $R_t$, which respectively award moving Forward towards an open direction, distance to Obstacles, keeping in Middle, saving Time (FOMT). The complete expression is presented by equation (3).}

\begin{equation}
R =\begin{cases}
R_c & \text{if collision}\\
R_r & \text{if reach open space}\\
c_1R_{f}+c_2R_{o}+c_3R_{m}+c_4R_{t} & \text{otherwise}
\end{cases}
\end{equation}
Where we have a negative reward for collision, and also a positive reward for reaching the open space. Notice that we consider if the robot reaches the open space (set as the end of the narrow space), an episode is done. To detect the open space is simple, once the sum of two scans, in the straight right and left direction, becomes bigger than a threshold, we consider the robot has reached the open space - the end of the narrow space. Otherwise, in other general steps within an episode, the reward function comprises mentioned four components, $R_f$, $R_o$, $R_m$, and $R_t$, named by FOMT as well.

\begin{equation}
R_{f} = \sum_{k=0} ^ {n_f} \alpha_2 ^{k}\times v\times (V_{obs}[f+k]+V_{obs}[f-k])
\end{equation}

The moving-forward reward component $R_f$ defined above aims to control the heading direction and urge the robot to move forward, where the heading control is implemented by awarding the most open direction, which means the obstacle along this direction is none or far away by laser range value. Specifically, $V_{obs}[f]$ denotes the observation of the ahead direction (on the mid-line of the robot), and $V_{obs}[f\pm k]$ denotes its $n_f$ neighboring observations, the reward of which will be larger with larger range values (longer distance to the obstacle), and function for mentioned heading control. 
 While the linear speed $v$ can urge the robot to move forward because without it the robot can stop or back up to keep a larger distance to the obstacle ahead to obtain a higher reward from $V_{obs}$. Correspondingly, $v$ can be 0 or negative to punish the stop or back up, hereby the robot would be encouraged to move forward with a positive $v$ to strike a higher score. Besides, discount factor $\alpha_2$ is to make the observation straight ahead more prominent than other neighboring tilted observations.

\begin{equation}
G = \left\{ [V_{obs}[i]- V_{range}[i] | i\in \mathbb{N}\right\}_{ascending}
\end{equation}
\begin{equation}
R_{o} = \sum_{k = 0}^{n_o} \alpha_1 ^{k}\times \log G[k]
\end{equation}

In equation (5), set $G$ denotes a vector that contains the gap value between each observation $V_{obs}[i]$ and its corresponding safe range $V_{range}[i]$ in ascending order. In equation (6), the obstacle-distance reward component $R_o$ uses a log function to punish the close distance using the sorted gap vector $G$, wherein a discount factor $\alpha_1$ is used to make the impact of closer distance more prominent.

\begin{equation}
R_{m} = -\sum_{k=0}^{n_m} \alpha_3^{k}\times |V_{obs}[r-k]-V_{obs}[l+k]|
\end{equation}

The keeping-in-middle reward component $R_m$, defined in equation (7), would help keep the robot in the middle of the obstacles around the robot's two sides. The difference between the observation in the straightly right direction $V_{obs}[r]$ and the observation in the left direction $V_{obs}[l]$, as well as the difference between their $n_m$ neighboring symmetric observations with respect to the robot's symmetry axis, are leveraged to evaluate the robot's performance of keeping in the middle. Where $\alpha_3$ serves as a factor to discount the reward from the neighboring observations. 

\begin{equation}
R_{t} = \alpha_4
\end{equation}

The last time-saving component $R_t$ is quite simple, where $\alpha_4$ is a negative constant to punish the time passing.

\begin{table}[htb!]
\caption{Parameter Library}
\label{table: table1}
\centering
\begin{tabular}{ll|lll}
\hline
 \textbf{Parameter} & \textbf{Value} &  \textbf{Parameter} &  \textbf{Value} &  \\ \hline
 $n_{scans}$ & 32 & $n_{f}$ & 5 &  \\
 $r$ & $0.095(m)$ & $n_{o}$ & 12 &  \\
 $R_c$ & -50 & $n_{m}$ & 5 &  \\
 $R_r$ & 50  & $\alpha_{1-3}$ & 0.9 &  \\
 $c_{1-4}$ & 1 &  $\alpha_4$ & -1  &  \\ \hline
\end{tabular}
\end{table}

\subsection{Reinforcement Learning Workflow}

The training process is operated under an OpenAI gym framework, where in each step, the robot with a current state $s_t$ outputs an action $a_t$, and waits for a step time interval to get new state $s_{t+1}$ and a reward $r_t$. Until ``done'', the number of steps will keep growing in an episode. In our settings, each time step last for $0.2s$, which means 5 actions will be generated per second. As Fig. \ref{workflow} shows, the DDPG method is given as an example to explain the workflow within a training episode. Wherein the input state $S_t$ is comprised of the action pair of the linear speed $ v\in[-0.6,0.6] (m/s)$ and the steering angle $w\in[-0.6,0.6] (rad)$, i.e., $(v,w)$, and the observation vector $V_{obs}$, which is derived from our proposed rectangular safety region and contains the laser range observation $O_1$ to $O_n \in [0, 6](m) $. The actor and critic networks are both composed of two fully linked layers of size 512, and output respectively the action pair $a_t$ and the $Q$ value. Then the robot in the simulation conducts $a_t$ and obtains a new state $S_{t+1}$ and a $done$ flag, which indicates whether the episode is ended by a collision or reaching the open space. Furthermore, they will be fed back to the networks and will update their parameters. While for discrete action settings (DQN), action pairs are formulated by $(v,w)\in\left\{-0.6, 0.6\right\}\times\left\{-0.6, 0, 0.6\right\}$.

\section{Experiments and Evaluations}

\begin{figure*}[h] 
\centering 
\includegraphics[width=0.90\textwidth]{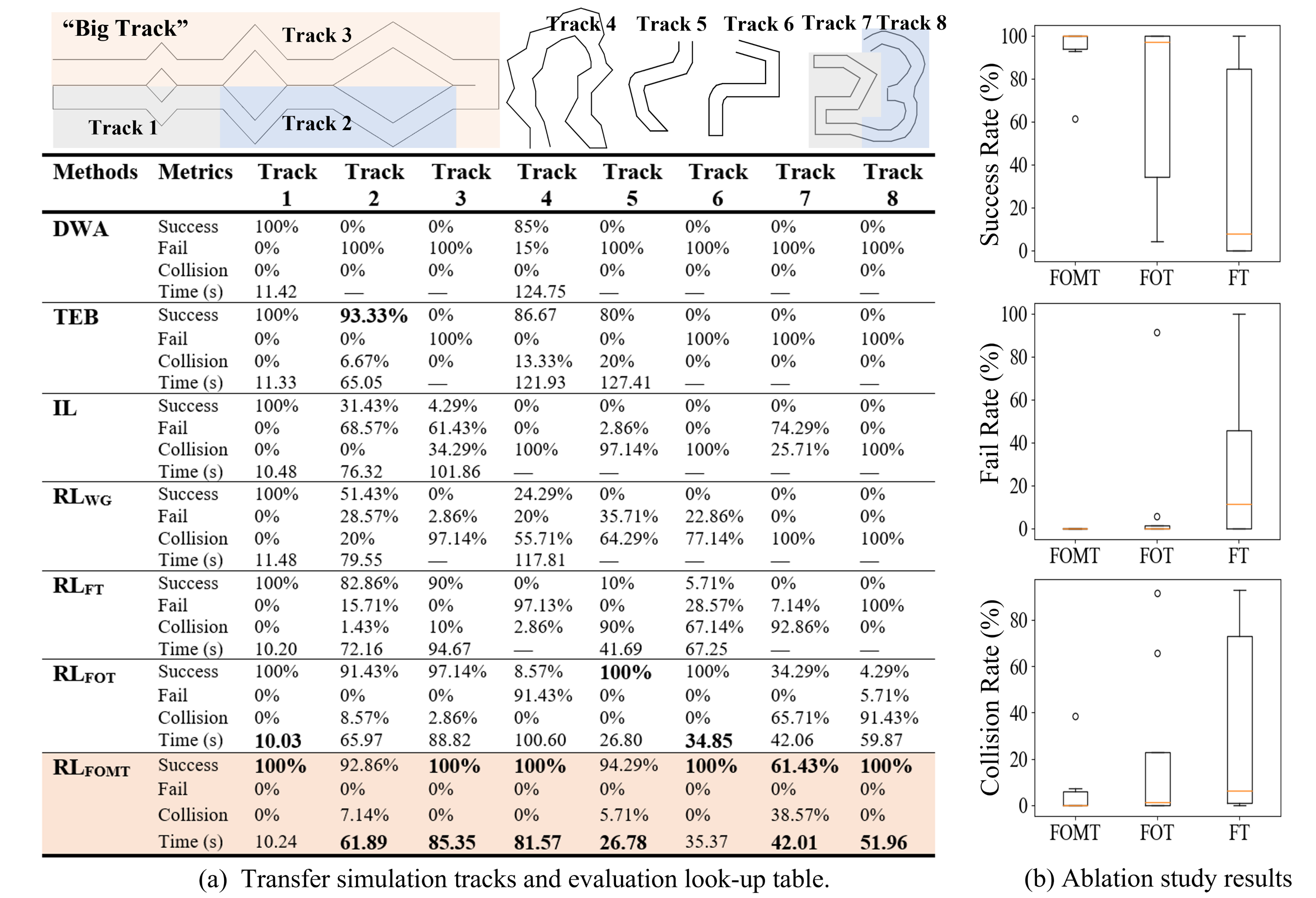} 
\caption{Evaluation results, quantitive
evaluation for all experiment sets and ablation studies.} 
\label{evaluation} 
\end{figure*}

Our experiments are conducted using the ZebraT platform~\cite{tian}, a rectangular-shaped robot with Lidar sensors, and a ROS/Gazebo simulator~\cite{tian} tailored for this robot. We conduct experiments in a simulated narrow ``big track" in the Gazebo simulator, which is a combination of Track 1, 2, and 3, as shown at the top of Figure~\ref{evaluation}. The "big track" contains 45, 90, and 180-degree corners, narrow corridors with a width of fewer than 2 meters, and the narrowest section is only around 1 meter wide, providing a rich training environment. The parameters used in the experiments are presented in Table~\ref{table: table1}. The contrastive and ablation study sets used in subsections B and C are explained in Table~\ref{table: table2}.

\begin{table}[htb!]
\caption{Experiment Library}
\label{table: table2}
\centering
\begin{tabular}{ll|lll}
\hline
 \textbf{Set} & \textbf{Note} &  \textbf{Set} &  \textbf{Note} &  \\ \hline
 $ \texttt{DWA} $ & map-based & $\texttt{TEB}$ & map-based &  \\
 $\texttt{IL}$ & imitation learning & $\texttt{RL}_{\texttt{WG}}$ &  waypoint-guided RL &  \\
 $\texttt{RL}_{\texttt{FT}}$ & RL with FT  & $\texttt{RL}_{\texttt{FOT}}$ & RL with FOT &  \\
$\texttt{RL}_{\texttt{FOMT}}$ & RL with FOMT  &  &  &  \\ \hline
\end{tabular}
\end{table}


\begin{figure}[h] 
\centering 
\includegraphics[width=0.95\columnwidth]{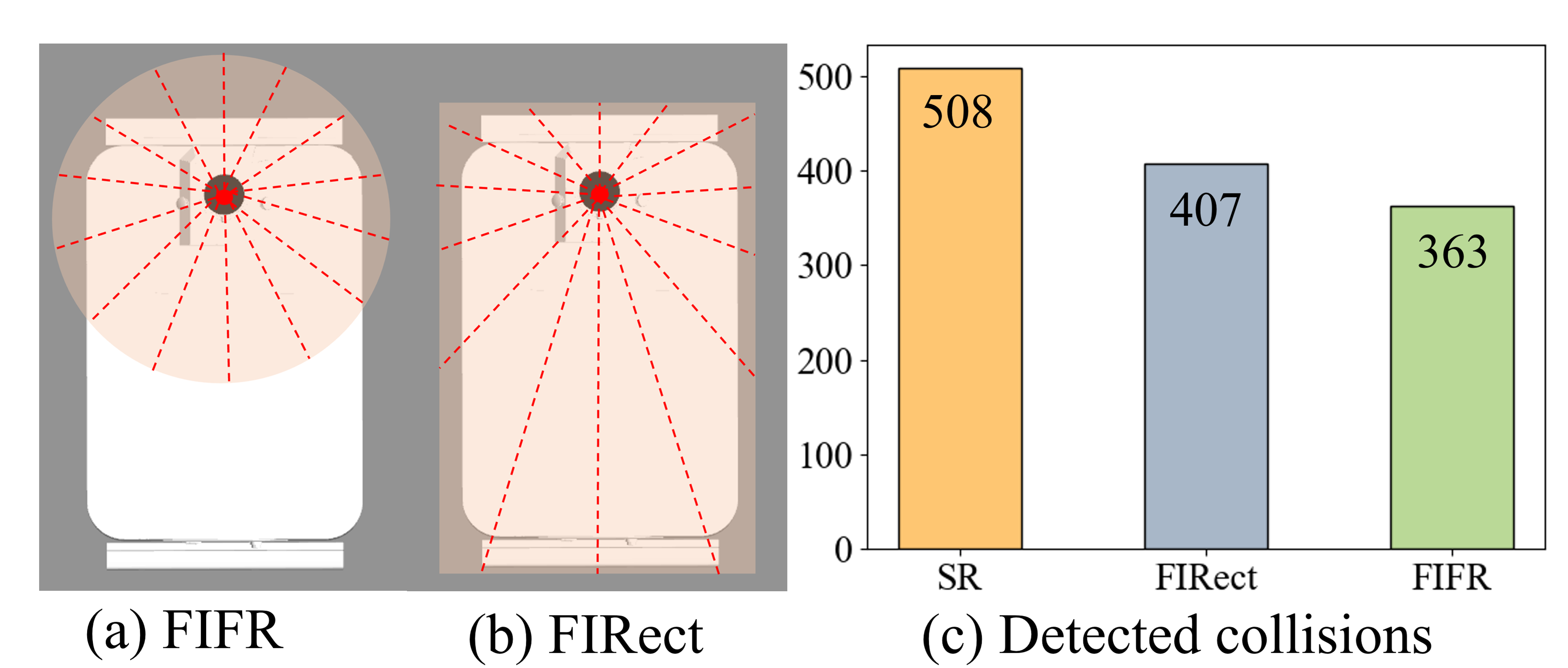} 
\caption{State representation comparison. (a) FIFR, fixed interval fixed range. (b) FIRect, fixed interval but different range to fit the rectangular contour. (c) Comparison by the number of detected collisions.} 
\label{collision} 
\end{figure}

\begin{figure}[h] %
\centering %
\includegraphics[width=0.75\columnwidth]{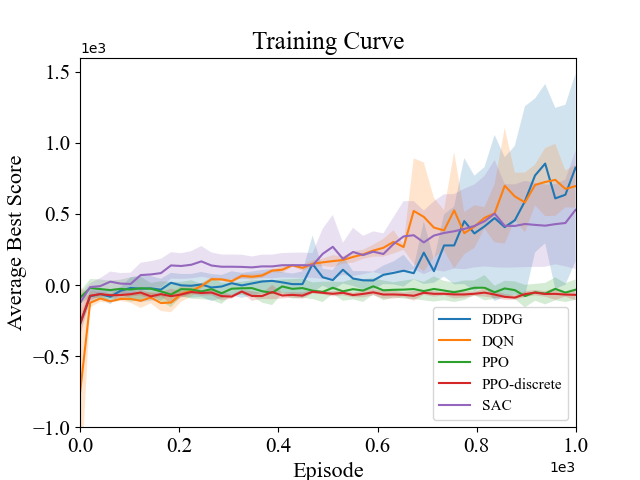} %
\caption{Training curve, the best score among every 20 episodes of five algorithms is plotted to show the training trend. Where the solid line denotes the average value of the best score of 5 random seeds and the shaded area denotes the trust region among random seeds.} %
\label{train} 
\end{figure}

\subsection{State Representation Comparison}

Collision detection is a critical aspect of operating in narrow spaces, as missing detections could lead to collisions and result in inappropriate learning by the DRL model. To validate our \textbf{Safe Region (SR)} state representation, we set up two baselines. Baseline \textbf{FIFR} uses Fixed Interval and Fixed Range (FIFR) to discretize laser scans, while Baseline \textbf{FIRect} uses Fixed Interval Rectangular-fitting (FIRect) ranges. FIFR is a popular representation paradigm used in previous works for round/near-round-shaped robots and serves as a contrastive set. FIRect serves as an ablation set to validate the interval discretization in our SR method. In Fig.~\ref{collision}(a, b), we present a visualization of FIFR and FIRect. To avoid over-coverage, we set the range value in FIFR to only cover the robot width. By operating robots in the ``big track" and conducting random collisions, we benchmarked the three paradigms by the number of detected collisions, as shown in Fig.~\ref{collision}(c). Our SR could detect most collisions, while FIRect and FIFR could detect $19.88\%$ and $28.54\%$ fewer collisions, respectively. This validates our SR state representation method.



\subsection{Training and Baseline Settings }

To select an appropriate DRL algorithm, five algorithms including DDPG, DQN, SAC, PPO, and PPO-discrete are benchmarked with our proposed state representation and reward function in the training process in the ``big track''. And the training curve is shown in Fig.~\ref{train}, where each algorithm is trained with five random seeds in 1000 episodes with memory buffer size at $2\times10^5$. The buffer size is designed based on an average number of sequences recorded in 1000 episodes, which varies from $3$ to $10\times10^5$ based on the total steps (a robot with better training performance would navigate longer, which takes more steps in an episode). DDPG could get the highest return and complete the track with learned backing-up collision-avoiding skills from our observation, thereby we select DDPG as the DRL algorithm in the subsequent experiments. And since the training rewards are the only factors to motivate robots for their training instead of final evaluation metrics, e.g., success rate and collision rate, to decrease noise and illustrate the training tendency, we pick the best score among every 20 episodes to formulate the training curve.

\begin{equation}
\label{eq:WG}
R =\begin{cases}
R_c & \text{if collision}\\
R_g & \text{if reach goal}\\
c_5(dist_{t-1}-dist_{t}) & \text{otherwise}
\end{cases}
\end{equation}

More importantly, we set up contrastive study and alation study to benchmark reward function shaping methods. In contrastive set \textbf{RL\textsubscript{WG}}, a traditional waypoint-guided reward function~\cite{trainning_wheels, discrete, moe} is defined in~\eqref{eq:WG} where the $R_c$ and $R_g$ are collision and goal rewards, which are set as -50 and 50 in the same scale as our reward function. In general steps, the distance difference to the waypoint formulates the reward with a coefficient $c_5$, which is set as 100 for scaling to the range of our reward function, and the waypoints are set manually along the tracks. Also, the relative distance between the robot pose and the goal pose, and the yaw angle difference are taken as additional 2 elements in our original safety region state vector of 32 laser range elements.

For ablation study, our set \textbf{RL\textsubscript{FOMT}} denotes that four components $R_f$,$R_o$,$R_m$,$R_t$ formulates our unguided reward function, whereas the ablation set \textbf{RL\textsubscript{FT}} and \textbf{RL\textsubscript{FOT}} are formulated respectively without $R_o$, $R_m$ and without $R_m$. The reason we do not ablate F or T component is that without F the robot will be not motivated to move, and T is just a general component that functions as a constant value. Moreover, the classical method sets \textbf{DWA}, \textbf{TEB}, and imitation learning set \textbf{IL} are formulated as well, and all the sets can be looked up in Table \ref{table: table2}. Specifically, to operate DWA and TEB, a premade map is provided, and local and global dilation radius is tuned with both 0.3 and 0.5 to obtain the best performance, and the robot wheelbase and steering radius are set up for TEB. Whereby the robot tends to encounter fewer planning failures but more collision risks with a 0.3 setting and tends to fail or oscillate with a 0.5 setting since the accessible area shrinks. While distance keeping ahead is set to 0.3 for TEB because a larger value could frequently cause failures in narrow space. In the IL training phase, a human expert operates the robot to drive in the ``big track'' for around 1 hour with 10000 states (laser range vectors) and actions recorded. Then a network with two 512 full-linked layers (similar to RL networks for fairness) is trained with state inputs with its action labels in a supervised paradigm. Besides, for each RL study set, training is conducted using 5 random seeds, 1000 training episodes with extra 500 episodes of best-model fine-tuning, and memory buffer size $2\times10^5$, while DDPG actor and critic-network learning rate at $1\times10^{-4}$ and $2\times10^{-4}$.

\subsection{Sim-to-sim Evaluation}
We conduct extensive evaluation experiments in 8 simulated narrow tracks, where Track 1-3 is from the training track, the ``big track'', and others are unseen tracks. Study sets listed in Table~\ref{table: table2} are compared by the metrics including success, fail, collision rate, and average time cost of success. For learning sets including IL, RL\textsubscript{WG}, RL\textsubscript{FT}, RL\textsubscript{FOT}, RL\textsubscript{FOMT}, all the best models are transferred to new tracks for evaluation and run 70 episodes. And for map-based sets, DWA and TEB, each setting of dilation radius 0.3 and 0.5 is respectively tested for 15 episodes, and only the well-performing setting (by success rate) is presented in the final result. \textbf{Quantitive evaluation} w.r.t. each method and each track are shown in Fig.~\ref{evaluation}(a), where our method RL\textsubscript{FOMT} strikes the highest success rate in most tracks with a lower time cost. 


\begin{figure*}[htbp] 
\centering 
\includegraphics[width=0.9\textwidth]{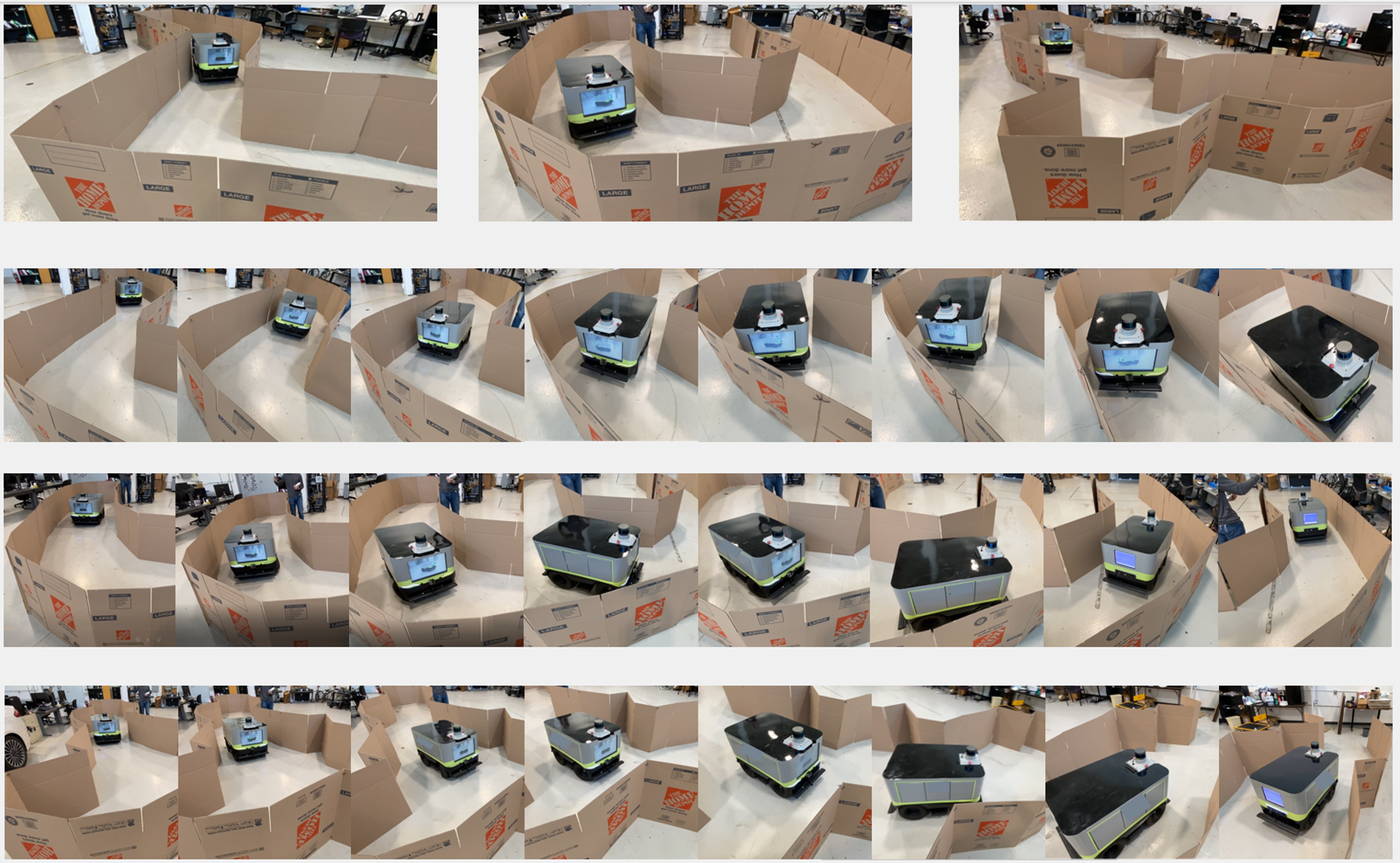} 
\caption{Real-world test, the first row shows Track 1 to 3 from left to right. And their demonstrations are respectively shown below. } 
\label{real} 
\end{figure*}

\textbf{Observations and analysis.} i) In classical map-based sets \textbf{DWA} and \textbf{TEB}, planning failures often happen in 90-degree turns, where the robot either oscillates or stays in place. Generally, TEB could outperform DWA since the algorithm per se considers car-like robot kinematics. While TEB enables some backing-up recovery actions but tends to reverse the robot's heading in some planning cases, which costs more time and is perceived as aberrant.
ii) For \textbf{IL} set, the model training requires high-quality demonstrations and time investments from human operators. We perceive significant difficulties for the IL model to be both validated in the seen scenario, i.e., ``big track'' and generalized to new scenarios. iii) For \textbf{RL\textsubscript{WG}} set using the traditional waypoint guided reward function, we perceive the method needs dense waypoints for training and the model could hardly master the 90-degree turning skills. We consider this could be induced by its ``collision-sparse'' reward function. Although it awards general steps and is not the traditional ``sparse reward function'' that only grades rewards for success or failure, we consider it grade collision in a sparse way so that only robot collides could make it learn the discrete collision boundary. Whereas in our FOMT reward function, the O component would grade on the distance to obstacles, which makes ``collision-sparse'' to ``collision-gradient''. Also, the ablation set RL\textsubscript{FT} is also considered ``collision-sparse'', hereby their collision rate is comparatively higher in Fig.~\ref{evaluation}(a). iv) Among the ablation sets \textbf{RL\textsubscript{FOMT}}, \textbf{RL\textsubscript{FT}}, and \textbf{RL\textsubscript{FOT}}, RL\textsubscript{FT} has the highest fail and collision rate shown in Fig.~\ref{evaluation}(b). By adding the O component, RL\textsubscript{FOT} achieves better collision avoidance performance than RL\textsubscript{FT}. And we perceive O component helps the robot to learn a backing-up skill to complete a 90-degree turn and thus RL\textsubscript{FOT}, and RL\textsubscript{FOMT} have higher success rates in Track 7 which contains continuous turns. However, RL\textsubscript{FT} and RL\textsubscript{FOT} both fail in Track 4 and 8 where the track is uneven or curvy, this unseen features impact their performance. While thanks to the M component, RL\textsubscript{FOMT} could motivate the robot to move in the middle of the side obstacles regardless of their features, hereby getting better performance.  In summary, O helps avoid a near collision and learn a continuous collision boundary, and M helps navigate among uneven or curvy obstacles in narrow spaces.

\subsection{Sim-to-real Demonstration}
Trained in the ``big track'', the well-performed RL\textsubscript{FOMT} model then is transferred to real-world narrow tracks shown in Fig.~\ref{real}. As a result, our ZebraT robot could complete all three tracks without any collisions. This demonstrates the scalability of the trained model and the validity of our design of reward function and related training works.

\section{Conclusion}
In this study, we propose a safety region-based representation paradigm tailored for the car-like rectangular-shaped robot and a FOMT-integrated reward function to explore narrow spaces without a map and waypoint guidance. By collision detection test among different state representation paradigms, we validate our claim on the effectiveness of the safety region. By intensive experiments in transferred seen and unseen tracks, we demonstrate the model with FOMT reward design could outperform the classical, IL methods and the RL method with guided reward design by contrastive studies. Besides, we analyzed the functionalities of the O and M component by ablation studies. Hereby we prove our proposed reward shaping is effective and interpretable. By final demonstration in the real-world narrow tracks, the policy generalizing power is validated.

\section*{Acknowledgment}

The work is partially supported by the US National Science Foundation (NSF) under Grant No. 2140346. The authors also wish to express their gratitude to all individuals and organizations that contributed directly or indirectly to the success of this work.

\bibliographystyle{IEEEtran}
\bibliography{main}

\begin{thebibliography}{10}
\providecommand{\url}[1]{#1}
\csname url@samestyle\endcsname
\providecommand{\newblock}{\relax}
\providecommand{\bibinfo}[2]{#2}
\providecommand{\BIBentrySTDinterwordspacing}{\spaceskip=0pt\relax}
\providecommand{\BIBentryALTinterwordstretchfactor}{4}
\providecommand{\BIBentryALTinterwordspacing}{\spaceskip=\fontdimen2\font plus
\BIBentryALTinterwordstretchfactor\fontdimen3\font minus
  \fontdimen4\font\relax}
\providecommand{\BIBforeignlanguage}[2]{{%
\expandafter\ifx\csname l@#1\endcsname\relax
\typeout{** WARNING: IEEEtran.bst: No hyphenation pattern has been}%
\typeout{** loaded for the language `#1'. Using the pattern for}%
\typeout{** the default language instead.}%
\else
\language=\csname l@#1\endcsname
\fi
#2}}
\providecommand{\BIBdecl}{\relax}
\BIBdecl

\bibitem{mining}
V.~Androulakis, J.~Sottile, S.~Schafrik, and Z.~Agioutantis, ``Concepts for
  development of autonomous coal mine shuttle cars,'' \emph{IEEE Transactions
  on Industry Applications}, vol.~56, no.~3, pp. 3272--3280, 2020.

\bibitem{post}
J.~Q. Cui, S.~K. Phang, K.~Z. Ang, F.~Wang, X.~Dong, Y.~Ke, S.~Lai, K.~Li,
  X.~Li, F.~Lin \emph{et~al.}, ``Drones for cooperative search and rescue in
  post-disaster situation,'' in \emph{2015 IEEE 7th international conference on
  cybernetics and intelligent systems (CIS) and IEEE conference on robotics,
  automation and mechatronics (RAM)}.\hskip 1em plus 0.5em minus 0.4em\relax
  IEEE, 2015, pp. 167--174.

\bibitem{libai1}
Y.~Ouyang, B.~Li, Y.~Zhang, T.~Acarman, Y.~Guo, and T.~Zhang, ``Fast and
  optimal trajectory planning for multiple vehicles in a nonconvex and
  cluttered environment: Benchmarks, methodology, and experiments,'' in
  \emph{2022 International Conference on Robotics and Automation (ICRA)}.\hskip
  1em plus 0.5em minus 0.4em\relax IEEE, 2022, pp. 10\,746--10\,752.

\bibitem{libai2}
C.~Sun, Q.~Li, B.~Li, and L.~Li, ``A successive linearization in feasible set
  algorithm for vehicle motion planning in unstructured and low-speed
  scenarios,'' \emph{IEEE Transactions on Intelligent Transportation Systems},
  vol.~23, no.~4, pp. 3724--3736, 2021.

\bibitem{DQN}
V.~Mnih, K.~Kavukcuoglu, D.~Silver, A.~Graves, I.~Antonoglou, D.~Wierstra, and
  M.~Riedmiller, ``Playing atari with deep reinforcement learning,''
  \emph{arXiv preprint arXiv:1312.5602}, 2013.

\bibitem{Go}
D.~Silver, A.~Huang, C.~J. Maddison, A.~Guez, L.~Sifre, G.~Van Den~Driessche,
  J.~Schrittwieser, I.~Antonoglou, V.~Panneershelvam, M.~Lanctot \emph{et~al.},
  ``Mastering the game of go with deep neural networks and tree search,''
  \emph{nature}, vol. 529, no. 7587, pp. 484--489, 2016.

\bibitem{arm}
K.~Nishi and N.~Nakasu, ``Evolvable motion-planning method using deep
  reinforcement learning,'' in \emph{2021 IEEE International Conference on
  Robotics and Automation (ICRA)}.\hskip 1em plus 0.5em minus 0.4em\relax IEEE,
  2021, pp. 4079--4085.

\bibitem{trainning_wheels}
L.~Xie, S.~Wang, S.~Rosa, A.~Markham, and N.~Trigoni, ``Learning with training
  wheels: speeding up training with a simple controller for deep reinforcement
  learning,'' in \emph{2018 IEEE International Conference on Robotics and
  Automation (ICRA)}.\hskip 1em plus 0.5em minus 0.4em\relax IEEE, 2018, pp.
  6276--6283.

\bibitem{its}
A.~Haydari and Y.~Yilmaz, ``Deep reinforcement learning for intelligent
  transportation systems: A survey,'' \emph{IEEE Transactions on Intelligent
  Transportation Systems}, 2020.

\bibitem{DDPG}
T.~P. Lillicrap, J.~J. Hunt, A.~Pritzel, N.~Heess, T.~Erez, Y.~Tassa,
  D.~Silver, and D.~Wierstra, ``Continuous control with deep reinforcement
  learning,'' \emph{arXiv preprint arXiv:1509.02971}, 2015.

\bibitem{SAC}
T.~Haarnoja, A.~Zhou, P.~Abbeel, and S.~Levine, ``Soft actor-critic: Off-policy
  maximum entropy deep reinforcement learning with a stochastic actor,'' in
  \emph{International conference on machine learning}.\hskip 1em plus 0.5em
  minus 0.4em\relax PMLR, 2018, pp. 1861--1870.

\bibitem{PPO}
J.~Schulman, F.~Wolski, P.~Dhariwal, A.~Radford, and O.~Klimov, ``Proximal
  policy optimization algorithms,'' \emph{arXiv preprint arXiv:1707.06347},
  2017.

\bibitem{astar}
P.~E. Hart, N.~J. Nilsson, and B.~Raphael, ``A formal basis for the heuristic
  determination of minimum cost paths,'' \emph{IEEE transactions on Systems
  Science and Cybernetics}, vol.~4, no.~2, pp. 100--107, 1968.

\bibitem{dijkstra}
E.~W. Dijkstra, ``A note on two problems in connexion with graphs,'' in
  \emph{Edsger Wybe Dijkstra: His Life, Work, and Legacy}, 2022, pp. 287--290.

\bibitem{ros}
M.~Quigley, K.~Conley, B.~Gerkey, J.~Faust, T.~Foote, J.~Leibs, R.~Wheeler,
  A.~Y. Ng \emph{et~al.}, ``Ros: an open-source robot operating system,'' in
  \emph{ICRA workshop on open source software}, vol.~3, no. 3.2.\hskip 1em plus
  0.5em minus 0.4em\relax Kobe, Japan, 2009, p.~5.

\bibitem{dwa}
D.~Fox, W.~Burgard, and S.~Thrun, ``The dynamic window approach to collision
  avoidance,'' \emph{IEEE Robotics \& Automation Magazine}, vol.~4, no.~1, pp.
  23--33, 1997.

\bibitem{teb}
C.~R{\"o}smann, W.~Feiten, T.~W{\"o}sch, F.~Hoffmann, and T.~Bertram,
  ``Trajectory modification considering dynamic constraints of autonomous
  robots,'' in \emph{ROBOTIK 2012; 7th German Conference on Robotics}.\hskip
  1em plus 0.5em minus 0.4em\relax VDE, 2012, pp. 1--6.

\bibitem{vo}
P.~Fiorini and Z.~Shiller, ``Motion planning in dynamic environments using
  velocity obstacles,'' \emph{The international journal of robotics research},
  vol.~17, no.~7, pp. 760--772, 1998.

\bibitem{rvo}
J.~Van~den Berg, M.~Lin, and D.~Manocha, ``Reciprocal velocity obstacles for
  real-time multi-agent navigation,'' in \emph{2008 IEEE international
  conference on robotics and automation}.\hskip 1em plus 0.5em minus
  0.4em\relax Ieee, 2008, pp. 1928--1935.

\bibitem{orca}
J.~Van Den~Berg, S.~J. Guy, M.~Lin, and D.~Manocha, ``Reciprocal n-body
  collision avoidance,'' in \emph{Robotics Research: The 14th International
  Symposium ISRR}.\hskip 1em plus 0.5em minus 0.4em\relax Springer, 2011, pp.
  3--19.

\bibitem{il}
M.~Pfeiffer, M.~Schaeuble, J.~Nieto, R.~Siegwart, and C.~Cadena, ``From
  perception to decision: A data-driven approach to end-to-end motion planning
  for autonomous ground robots,'' in \emph{2017 ieee international conference
  on robotics and automation (icra)}.\hskip 1em plus 0.5em minus 0.4em\relax
  IEEE, 2017, pp. 1527--1533.

\bibitem{curiosity}
O.~Zhelo, J.~Zhang, L.~Tai, M.~Liu, and W.~Burgard, ``Curiosity-driven
  exploration for mapless navigation with deep reinforcement learning,''
  \emph{arXiv preprint arXiv:1804.00456}, 2018.

\bibitem{discrete}
E.~Marchesini and A.~Farinelli, ``Discrete deep reinforcement learning for
  mapless navigation,'' in \emph{2020 IEEE International Conference on Robotics
  and Automation (ICRA)}.\hskip 1em plus 0.5em minus 0.4em\relax IEEE, 2020,
  pp. 10\,688--10\,694.

\bibitem{autorl}
H.-T.~L. Chiang, A.~Faust, M.~Fiser, and A.~Francis, ``Learning navigation
  behaviors end-to-end with autorl,'' \emph{IEEE Robotics and Automation
  Letters}, vol.~4, no.~2, pp. 2007--2014, 2019.

\bibitem{unknown_dynamic}
J.~Zeng, R.~Ju, L.~Qin, Y.~Hu, Q.~Yin, and C.~Hu, ``Navigation in unknown
  dynamic environments based on deep reinforcement learning,'' \emph{Sensors},
  vol.~19, no.~18, p. 3837, 2019.

\bibitem{robust}
F.~Leiva and J.~Ruiz-del Solar, ``Robust rl-based map-less local planning:
  Using 2d point clouds as observations,'' \emph{IEEE Robotics and Automation
  Letters}, vol.~5, no.~4, pp. 5787--5794, 2020.

\bibitem{motion_dynamic}
M.~Everett, Y.~F. Chen, and J.~P. How, ``Motion planning among dynamic,
  decision-making agents with deep reinforcement learning,'' in \emph{2018
  IEEE/RSJ International Conference on Intelligent Robots and Systems
  (IROS)}.\hskip 1em plus 0.5em minus 0.4em\relax IEEE, 2018, pp. 3052--3059.

\bibitem{DWA_RL}
U.~Patel, N.~K.~S. Kumar, A.~J. Sathyamoorthy, and D.~Manocha, ``Dwa-rl:
  Dynamically feasible deep reinforcement learning policy for robot navigation
  among mobile obstacles,'' in \emph{2021 IEEE International Conference on
  Robotics and Automation (ICRA)}.\hskip 1em plus 0.5em minus 0.4em\relax IEEE,
  2021, pp. 6057--6063.

\bibitem{moe}
W.~Zhang, K.~Zhao, P.~Li, X.~Zhu, F.~Ye, W.~Jiang, H.~Fu, and T.~Wang,
  ``Learning to navigate in a vuca environment: Hierarchical multi-expert
  approach,'' in \emph{2021 IEEE/RSJ International Conference on Intelligent
  Robots and Systems (IROS)}.\hskip 1em plus 0.5em minus 0.4em\relax IEEE,
  2021, pp. 9254--9261.

\bibitem{mover}
J.~H. Reif, ``Complexity of the mover's problem and generalizations,'' in
  \emph{20th Annual Symposium on Foundations of Computer Science (sfcs
  1979)}.\hskip 1em plus 0.5em minus 0.4em\relax IEEE Computer Society, 1979,
  pp. 421--427.

\bibitem{narrow}
P.~Polack, L.-M. Dallen, and A.~Cord, ``Strategy for automated dense parking:
  how to navigate in narrow lanes,'' in \emph{2020 IEEE International
  Conference on Robotics and Automation (ICRA)}.\hskip 1em plus 0.5em minus
  0.4em\relax IEEE, 2020, pp. 9196--9202.

\bibitem{bnd}
K.~Wu, H.~Wang, M.~A. Esfahani, and S.~Yuan, ``Bnd*-ddqn: Learn to steer
  autonomously through deep reinforcement learning,'' \emph{IEEE Transactions
  on Cognitive and Developmental Systems}, vol.~13, no.~2, pp. 249--261, 2019.

\bibitem{successor_features}
J.~Zhang, J.~T. Springenberg, J.~Boedecker, and W.~Burgard, ``Deep
  reinforcement learning with successor features for navigation across similar
  environments,'' in \emph{2017 IEEE/RSJ International Conference on
  Intelligent Robots and Systems (IROS)}.\hskip 1em plus 0.5em minus
  0.4em\relax IEEE, 2017, pp. 2371--2378.

\bibitem{target}
Y.~Zhu, R.~Mottaghi, E.~Kolve, J.~J. Lim, A.~Gupta, L.~Fei-Fei, and A.~Farhadi,
  ``Target-driven visual navigation in indoor scenes using deep reinforcement
  learning,'' in \emph{2017 IEEE international conference on robotics and
  automation (ICRA)}.\hskip 1em plus 0.5em minus 0.4em\relax IEEE, 2017, pp.
  3357--3364.

\bibitem{towards_cognitive}
L.~Tai and M.~Liu, ``Towards cognitive exploration through deep reinforcement
  learning for mobile robots,'' \emph{arXiv preprint arXiv:1610.01733}, 2016.

\bibitem{towards_monocular}
L.~Xie, S.~Wang, A.~Markham, and N.~Trigoni, ``Towards monocular vision based
  obstacle avoidance through deep reinforcement learning,'' \emph{arXiv
  preprint arXiv:1706.09829}, 2017.

\bibitem{virtual_to_real}
L.~Tai, G.~Paolo, and M.~Liu, ``Virtual-to-real deep reinforcement learning:
  Continuous control of mobile robots for mapless navigation,'' in \emph{2017
  IEEE/RSJ International Conference on Intelligent Robots and Systems
  (IROS)}.\hskip 1em plus 0.5em minus 0.4em\relax IEEE, 2017, pp. 31--36.

\bibitem{AC}
V.~Konda and J.~Tsitsiklis, ``Actor-critic algorithms,'' \emph{Advances in
  neural information processing systems}, vol.~12, 1999.

\bibitem{PG}
R.~S. Sutton, D.~McAllester, S.~Singh, and Y.~Mansour, ``Policy gradient
  methods for reinforcement learning with function approximation,''
  \emph{Advances in neural information processing systems}, vol.~12, 1999.

\bibitem{tian}
Z.~Tian and W.~Shi, ``Design and implement an enhanced simulator for autonomous
  delivery robot,'' in \emph{2022 Fifth International Conference on Connected
  and Autonomous Driving (MetroCAD)}, 2022, pp. 21--29.

\end{thebibliography}

\end{document}